%% file: acl_latex.tex
\pdfoutput=1
\documentclass[11pt]{article}
\usepackage{acl}
\usepackage{times}
\usepackage{latexsym}
\usepackage[T1]{fontenc}
\usepackage[utf8]{inputenc}
\usepackage{microtype}
\usepackage{inconsolata}
\usepackage{graphicx}
\usepackage{booktabs}
\usepackage{xcolor}
\usepackage{makecell}
\usepackage{placeins}
\usepackage{svg}
\usepackage{amsmath}
\usepackage{makecell}
\usepackage{multirow}
\usepackage{adjustbox}
\usepackage{subcaption}
\usepackage{caption}
\usepackage{float}
\usepackage{graphicx}
\usepackage{svg}
\usepackage{enumitem}
\svgpath{{figures/plots/pipeline/}}

\title{\hspace{-20pt}\raisebox{-0.35\height}{\includegraphics[width=.0999\textwidth]{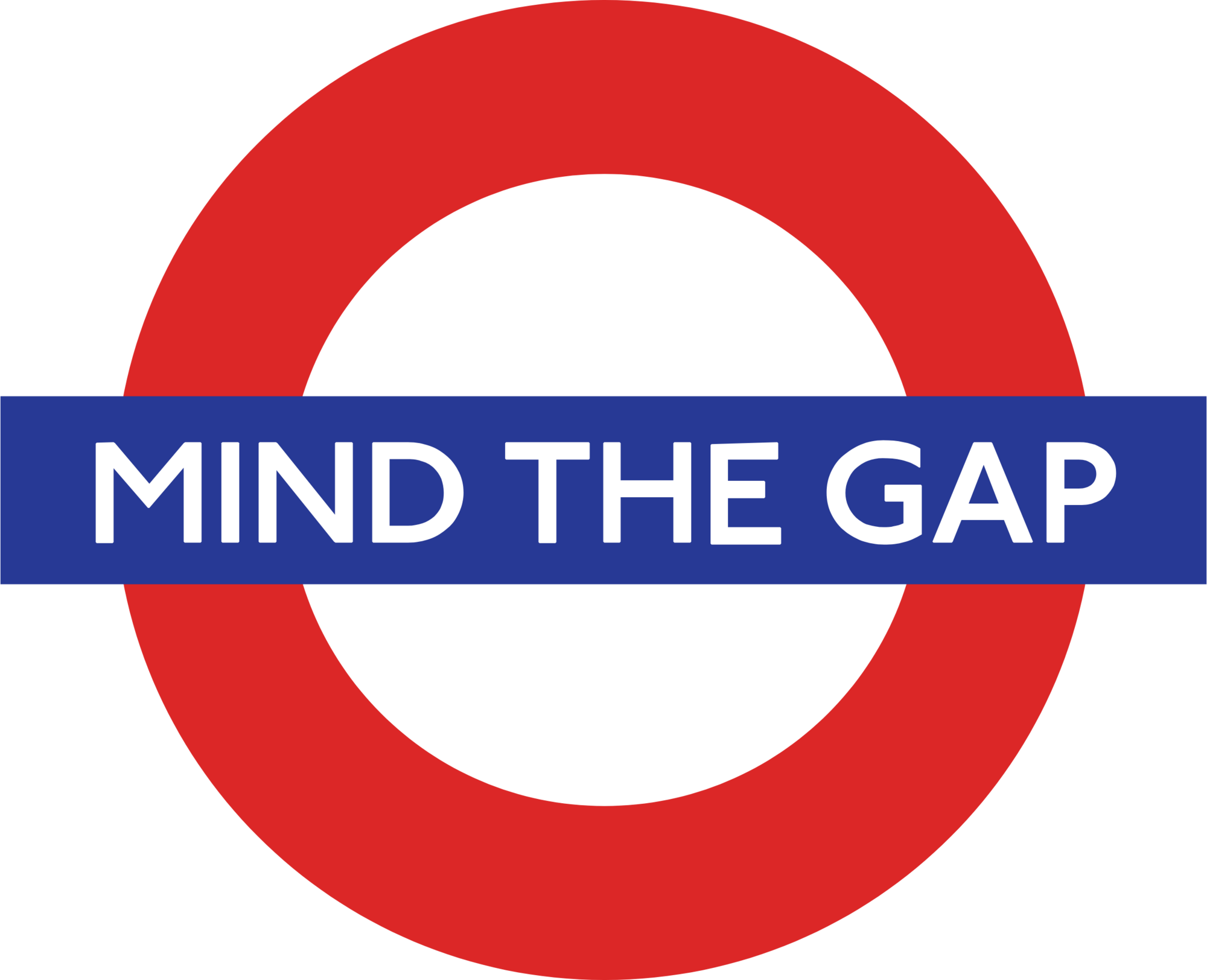}}\hspace{3pt} A Review of Arabic Post-Training Datasets and Their Limitations}

\input{macros}

\author{
Mohammed Alkhowaiter$^{1}$\thanks{Contributed equally; contributions varied by focus.}\thanks{Corresponding author: \texttt{\href{mailto:mohammed@refineai.dev}{mohammed@refineai.dev}}.} \quad 
Norah Alshahrani$^{2,*}$ \quad 
Saied Alshahrani$^{3,*}$ \\
\textbf{Reem I. Masoud}$^{4,7,*}$ \quad  
\textbf{Alaa Alzahrani}$^{5,*}$ \quad 
\textbf{Deema Alnuhait}$^{6,*}$ \\
\textbf{Emad A. Alghamdi}$^{8}$ \quad  
\textbf{Khalid Almubarak}$^{8}$ \\
$^{1}$Refine AI \quad $^{2}$ASAS AI \quad
$^{3}$University of Bisha\\
$^{4}$University College London \quad
$^{5}$King Salman Global Academy for Arabic \\
$^{6}$University of Illinois at Urbana-Champaign \quad
$^{7}$King Abdulaziz University\\
$^{8}$HUMAIN\\[2ex]
}

\begin{document}
\maketitle

\begin{abstract} 
Post-training has emerged as a crucial technique for aligning pre-trained Large Language Models (LLMs) with human instructions, significantly enhancing their performance across a wide range of tasks. Central to this process is the quality and diversity of post-training datasets. This paper presents a review of publicly available Arabic post-training datasets on the Hugging Face Hub, organized along four key dimensions: (1) LLM Capabilities (e.g., Question Answering, Translation, Reasoning, Summarization, Dialogue, Code Generation, and Function Calling); (2) Steerability (e.g., Persona and System Prompts); (3) Alignment (e.g., Cultural, Safety, Ethics, and Fairness); and (4) Robustness. Each dataset is rigorously evaluated based on popularity, practical adoption, recency and maintenance, documentation and annotation quality, licensing transparency, and scientific contribution. Our review revealed critical gaps in the development of Arabic post-training datasets, including limited task diversity, inconsistent or missing documentation and annotation, and low adoption across the community. Finally, the paper discusses the implications of these gaps on the progress of Arabic-centric LLMs and applications while providing concrete recommendations for future efforts in Arabic post-training dataset development.
\end{abstract}

\begin{figure}[ht]
    \centering
    \includegraphics[width=0.5\textwidth]{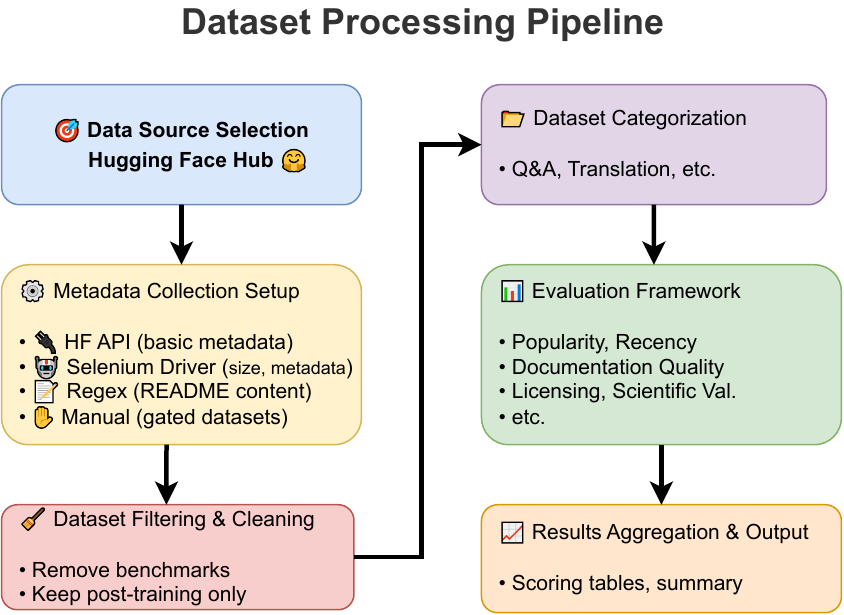}
    \caption{General Processing Pipeline for Arabic Post-Training Dataset Collection, Filtering, and Evaluation.}
    \label{fig:method_pipeline}
\end{figure}

\section{Introduction}

Recent years there has been a growing interest in building high-quality post-training datasets to steer and enhance the capabilities of Large Language Models (LLMs). The nature of post-training has evolved alongside advancements in AI models. Although post-training still occurs after pre-training on large text corpora, its focus has shifted. Previously, post-training often involved task-specific fine-tuning, such as sentiment analysis, topic classification, or image classification, with models like BERT \cite{devlin2019bertpretrainingdeepbidirectional}. Today, it has expanded into a broader and more general concept.

This shift became clear with the emergence of capabilities of LLMs, as highlighted by \citet{brown2020languagemodelsfewshotlearners}, which demonstrated strong performance on various tasks through zero-shot or few-shot prompting, even without explicit task-specific training. These capabilities were further advanced by works like \citet{ouyang2022traininglanguagemodelsfollow}, which aligned models to better follow user intent, enabling more engaging and coherent interactions in dialogue formats to utilize these capabilities. This trend has also extended to other languages, such as Arabic, which has witnessed significant growth through several Arabic-centric LLMs, aimed at enhancing and improving post-training datasets.

A variety of strategies have been utilized to develop post-training datasets tailored to Arabic-centric LLMs. For the JAIS models \cite{sengupta2023jaisjaischatarabiccentricfoundation}, instruction tuning was performed using a mix of English and Arabic datasets. The Arabic portion was primarily composed of translated adaptations of widely adopted English post-training resources, including those from \citet{wang2022supernaturalinstructionsgeneralizationdeclarativeinstructions, alpaca, DatabricksBlog2023DollyV2}, along with template-based instruction datasets such as \citet{muennighoff2022crosslingual}. In addition to these translated datasets, two original datasets—NativeQA-Ar and SafetyQA-Ar—were specifically developed to incorporate culturally and contextually relevant content for the United Arab Emirates and the wider Arab region.

\citet{huang2024acegptlocalizinglargelanguage}
introduced an Arabic-centric LLMs, dubbed AceGPT, by continuing pre-training from Llama 2~\cite{touvron2023llama}. In the post-training phase, their primary focus was on localizing instructions and preference data. They generated synthetic Arabic data by prompting GPT-4 model directly in Arabic, which resulted in more culturally nuanced responses compared to prompts in English. Additionally, they incorporated well-known datasets, such as Alpaca, Evol-Instruct, and Code-Alpaca, into their Supervised Fine-tuning (SFT) mixture and generated corresponding Arabic versions using GPT-4~\cite{achiam2023gpt}. ALLaM series of models \cite{bari2024allamlargelanguagemodels} were post-trained on datasets collected  from public and proprietary sources, covering a diverse range of topics, including education, history, Arabic linguistics, politics, and religion. Additionally, their post-training dataset underwent multiple filtering steps to ensure high quality. A more recent methodology proposed by \citet{fanar} introduced a synthetic data generation pipeline aimed at enriching post-training datasets with culturally contextualized content. Despite these significant efforts, publicly available Arabic post-training datasets remain considerably behind those of many other languages. Even the Arabic-centric LLMs developed to date still struggle to compete closely with known LLMs, whether open-source ones, like DeepSeek and Qwen, or proprietary models, like ChatGPT, Claude, and Gemini, according to the Open Arabic LLM Leaderboard by \citet{OALL-2}. 


A key reason behind this gap is that Arabic still underrepresented in post-training efforts~\cite{guellil2021arabic} even though it is a native language of over 400 million speakers across 22 countries, and its position as the fourth most used language on the Internet~\cite{boudad2018sentiment}. This underrepresentation is largely due to limited publication of Arabic post-training dataset. Moreover, the Arabic language has rich morphology, non-concatenative word formation, complex syntactic structures, and significant diglossia between Classical Arabic (CA),  Modern Standard Arabic (MSA), and Dialectal Arabic (DA), which introduce additional layers of ambiguity \cite{darwish-2014-arabizi}. Given Arabic’s linguistic complexity, cultural richness, and global relevance~\cite{bakalla2023arabic, versteegh2014arabic}, it is essential to rethink how post-training resources are developed for the language.

This paper surveys existing Arabic post-training datasets, identifies critical gaps, addresses  challenges, and offers recommendations, all to guide future Arabic post-training dataset development. We list our  main contributions as the following:
 \begin{itemize}
     \item We systematically reviewed publicly open Arabic datasets used for post-training and alignment of Arabic-centric language models.
     \item We developed  tools\footnote{\url{www.github.com/refineaidev/mind-the-gap}.} to automatically extract Arabic post-training datasets from the Hugging Face Hub and evaluate each dataset across six dimensions: documentation, popularity, adoption, recency and maintenance, licensing transparency, and scientific value.
     
     \item We identified critical gaps in Arabic post-training dataset development and offered recommendations to improve transparency, cultural relevance, and downstream usability.
 \end{itemize}

\section{Methodology}

We exclusively collected Arabic post-training datasets' metadata from the \textit{Hugging Face Hub}, as it represents the most comprehensive and widely-adopted machine learning platform utilized by researchers, developers, and organizations worldwide. 
While we initially attempted to diversify our sources by including platforms such as \textit{GitHub} and \textit{Kaggle}, the number of datasets with sufficient metadata and standardized formatting was negligible compared to Hugging Face Hub's extensive collection. Additionally, GitHub and Kaggle datasets often lack the structured metadata tags and consistent documentation standards essential for our automated collection methodology. Therefore, we focused solely on the \textit{Hugging Face Hub} as our primary source to ensure data quality, consistency, and comprehensive coverage of available Arabic post-training datasets. Our dataset collection and evaluation pipeline is shown in Figure~\ref{fig:method_pipeline}.

\subsection{Experimental Setup}
We utilized the Hugging Face Hub Python library to automatically collect the following metadata for each dataset: \textit{Dataset ID (dataset name)}, \textit{Number of Likes}, \textit{Number of Downloads}, \textit{Last Modified Date}, \textit{Name of License}, \textit{ArXiv Papers}, and \textit{Number of Models that have used this dataset}. We further employed the Selenium Python library to automate the collection of additional metadata not provided by the Hugging Face Hub Python library, including \textit{Size of Downloaded Files}, \textit{Size of Parquet Files}, and \textit{Number of Rows}.

\subsection{Metadata Collection}
We employed four distinct approaches to gather metadata for Arabic post-training datasets: 1) \textit{automatic collection} of metadata using the Hugging Face Hub Python library, leveraging the platform's metadata tags; 2) \textit{automated collection} of metadata using the Selenium Python library, extracting information from the dataset's statistics widget (located on the right side of the dataset card); 3) \textit{regular expression search} for specific metadata within README.md files of datasets, such as \textit{ACL Papers}, again utilizing the Hugging Face Hub Python library; and 4) \textit{manual collection} of metadata for gated datasets, which are private datasets requiring access requests, making automatic and automated collection approaches infeasible. We also manually removed benchmark datasets to ensure our collection exclusively contained post-training datasets.

\subsection{Evaluations of Datasets}

We evaluated Arabic post-training datasets across 12 task categories, mapped to four dimensions: (1) LLM Capabilities (e.g., \textit{Q\&A}, \textit{Translation}, \textit{Reasoning and Multi-Step Thinking}, \textit{Summarization}, \textit{Dialogue}, \textit{Code Generation}, and \textit{Function Calling}); (2) Steerability (e.g., \textit{Persona} and \textit{System Prompt}); (3) Alignment (e.g., \textit{Cultural Alignment}, \textit{Safety, Ethics, and Fairness}); and (4) Robustness. The selection of the 12 task categories was informed by two criteria: (1) alignment with established taxonomies in prior research, like \citet{chen2024mastering, minaee2024large}, and (2) representation of distinct, functionally coherent areas relevant to LLM evaluation and dataset availability. Specifically, we synthesized insights from \citet{minaee2024large}, who provide a broad survey of LLM capabilities across general NLP domains. This combined perspective ensured that our categories address both specialized applications, such as \textit{Code Generation}, and general-purpose tasks, such as \textit{Summarization}.

Each dataset was assessed using framework comprising six evaluation criteria: documentation and annotation quality, popularity, practical adoption, recency and active maintenance, licensing transparency, and scientific contribution. Each  criterion utilizes a structured scoring system designed for simplicity, consistency, and reproducibility.

To illustrate our methodology, Table~\ref{tab:tableA} presents an example of evaluation criteria and scoring rubrics used to assess documentation and annotation quality across datasets. We deliberately employed straightforward rubrics to ensure simplicity, efficiency, and effectiveness in our evaluation process. The remaining set of evaluation criteria and corresponding scoring systems for all assessment dimensions is provided in Appendix~\ref{appendix. criteria} (Table~\ref{tab:scoring-rubric}), offering full transparency in our methodology and enabling reproducibility of our findings.

\begin{table*}[!ht]
  \centering\small
  \caption{An detailed example of the evaluation criteria and scoring system used for evaluating documentation and annotation quality. The remaining evaluation criteria and scoring rubrics are provided in Appendix~\ref{appendix. criteria} (Table~\ref{tab:scoring-rubric}).}
  \begin{adjustbox}{width=\textwidth}
    \begin{tabular}{@{}l p{0.4\textwidth} c c@{}}
      \toprule
      \textbf{Evaluation} 
        & \textbf{Criteria} 
        & \textbf{Avg.\ Score} 
        & \textbf{Quality Level} \\
      \midrule

      \multirow{3}{*}{Documentation}
        & \textbullet\ Dataset card explains the usage of dataset\newline
          \textbullet\ Dataset card states the license clearly\newline
          \textbullet\ Dataset card shows examples of dataset\newline
          \textbullet\ Dataset card includes or cites a paper\newline
          \textbullet\ Dataset card describes the datasets\newline
          \textbullet\ Dataset card states the authors or maintainers
        & \multirow{3}{*}{\begin{tabular}{@{}c@{}}
            $4 \le \text{score} \le 6$\\
            \\
            $2 \le \text{score} < 4$\\
            \\
            $\text{score} < 2$
          \end{tabular}
          }
        & \multirow{3}{*}{\begin{tabular}{@{}c@{}}
            High\\
            \\
            Medium\\
            \\
            Low
          \end{tabular}} \\
      \midrule

      \multirow{3}{*}{Annotation}
        & \textbullet\ Metadata tags specify a task\newline
          \textbullet\ Metadata tags specify a language\newline
          \textbullet\ Metadata tags state a size\newline
          \textbullet\ Metadata tags state a license\newline
          \textbullet\ Metadata tags include dataset source\newline
          \textbullet\ Metadata tags include configurations
        & \multirow{3}{*}{\begin{tabular}{@{}c@{}}
            $4 \le \text{score} \le 6$\\
            \\
            $2 \le \text{score} < 4$\\
            \\
            $\text{score} < 2$
          \end{tabular}}
        & \multirow{3}{*}{\begin{tabular}{@{}c@{}}
            High\\
            \\
            Medium\\
            \\
            Low
          \end{tabular}} \\
      \bottomrule
    \end{tabular}
  \end{adjustbox}
  \label{tab:tableA}
  \vspace{-5pt}
\end{table*}

\section{Analysis and Results}

We analyzed 366 datasets across 12 Natural Language Processing (NLP) domains, summarized in Table~\ref{tab:dataset-summary}. Due to unbalanced group sizes and small sample sizes in certain domain categories, we present only descriptive statistics to avoid Type I and Type II errors associated with insufficient statistical power and unequal groups ~\cite{field2018discovering}. The remainder of this section will first cover the descriptive statistics of the collected datasets, followed by the evaluation results for those datasets.

\begin{figure}[ht]
    \centering
    \includegraphics[width=0.485\textwidth]{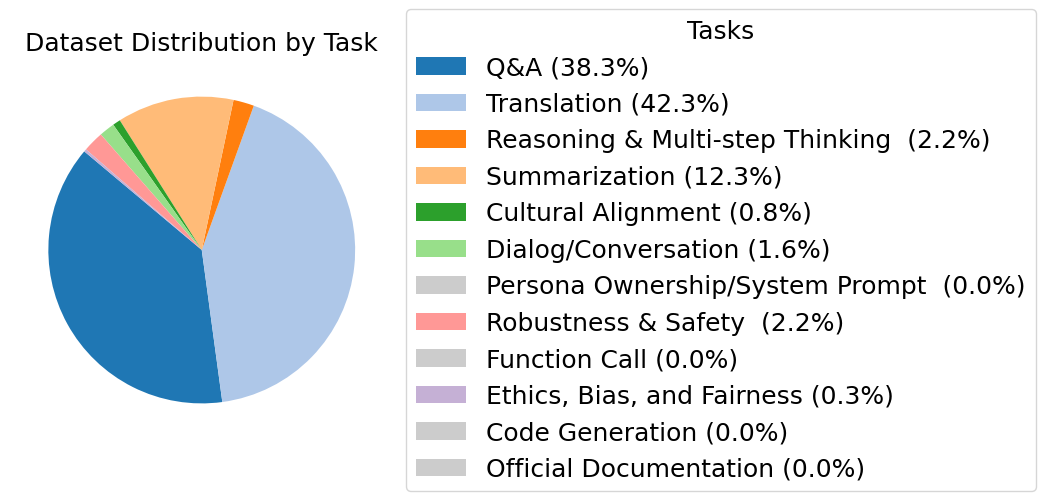}
    \caption{Distribution of datasets across tasks. Labels include the percentage of datasets in each task. Tasks with no datasets are shown for the sake of completeness.}
    \label{fig:dataset_distribution}

    \vspace{-5pt}
\end{figure}

\begin{figure*}[ht]
\centering

\begin{subfigure}[b]{0.32\textwidth}
    \includegraphics[width=\linewidth]{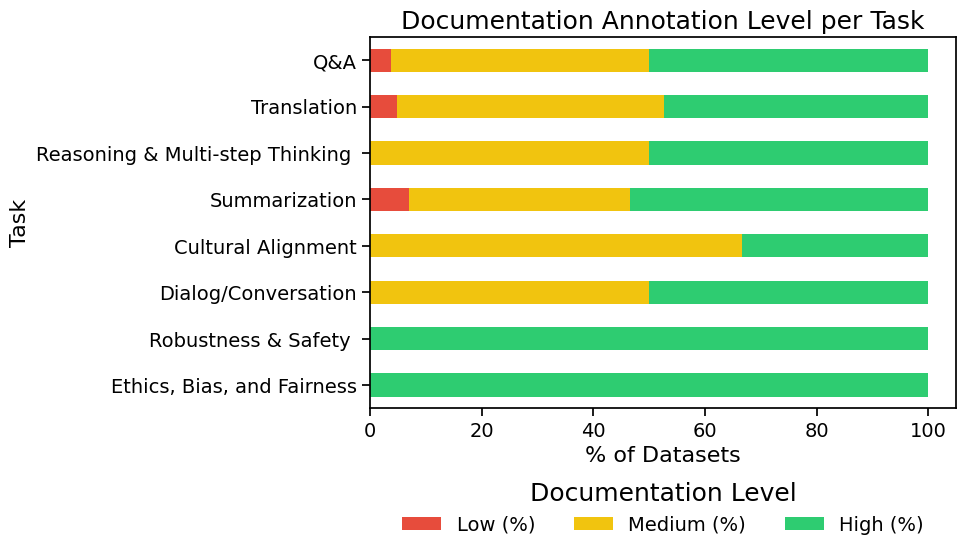}
    \caption{Documentation level distribution}
    \label{fig:doc_dist}
\end{subfigure}
\hfill
\begin{subfigure}[b]{0.32\textwidth}
    \includegraphics[width=\linewidth]{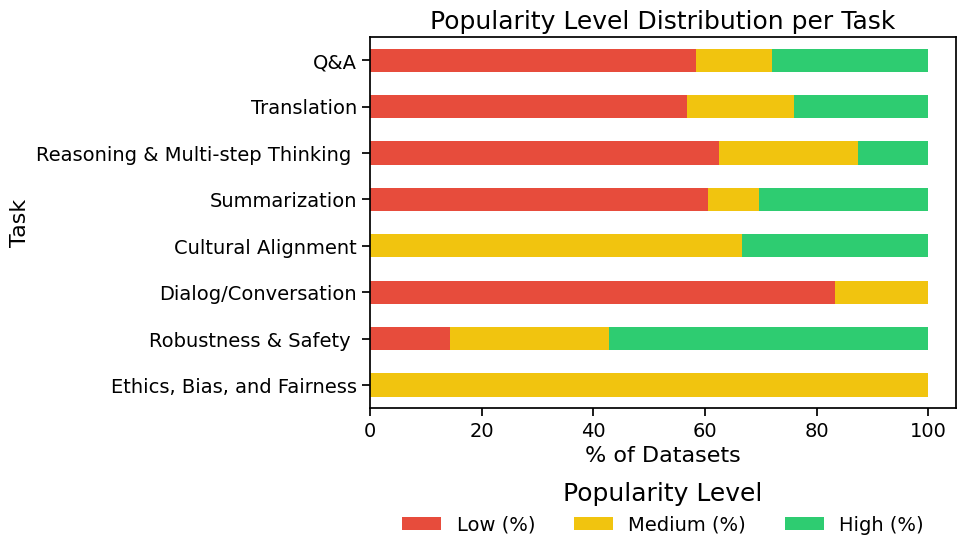}
    \caption{Popularity level distribution}
    \label{fig:popularity}
\end{subfigure}
\hfill
\begin{subfigure}[b]{0.32\textwidth}
    \includegraphics[width=\linewidth]{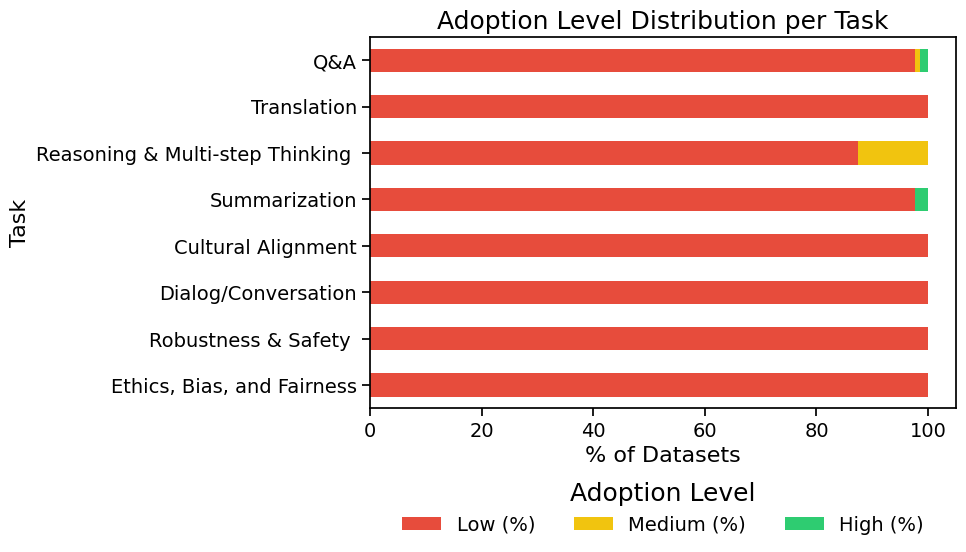}
    \caption{Adoption level distribution}
    \label{fig:adoption}
\end{subfigure}

\vspace{0.8em}

\begin{subfigure}[b]{0.32\textwidth}
    \includegraphics[width=\linewidth]{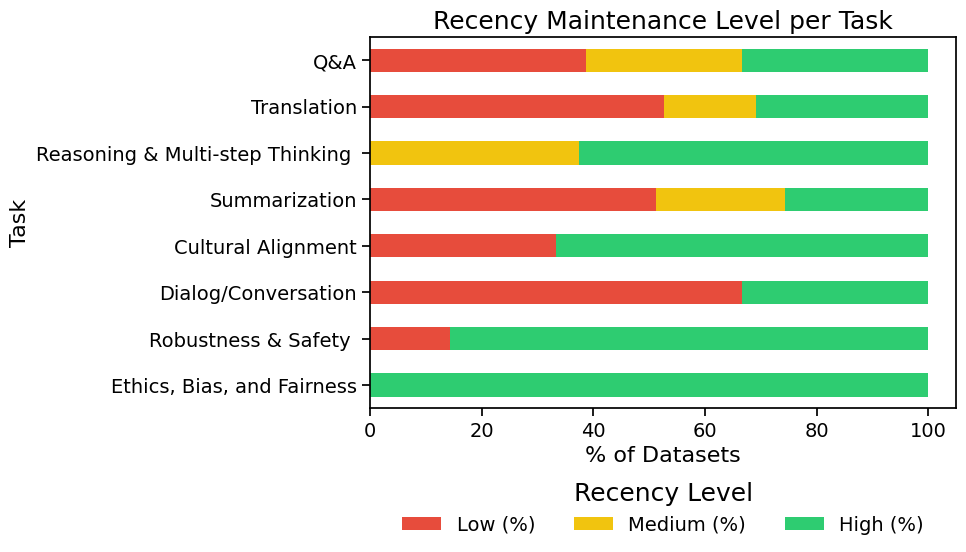}
    \caption{Recency maintenance per task}
    \label{fig:recency}
\end{subfigure}
\hfill
\begin{subfigure}[b]{0.32\textwidth}
    \includegraphics[width=\linewidth]{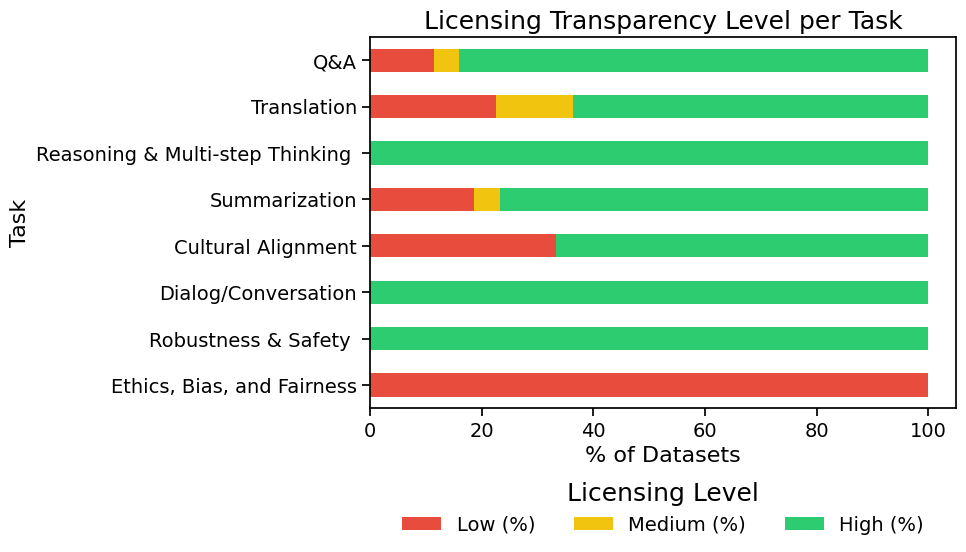}
    \caption{Licensing transparency per task}
    \label{fig:licencing}
\end{subfigure}
\hfill
\begin{subfigure}[b]{0.32\textwidth}
    \includegraphics[width=\linewidth]{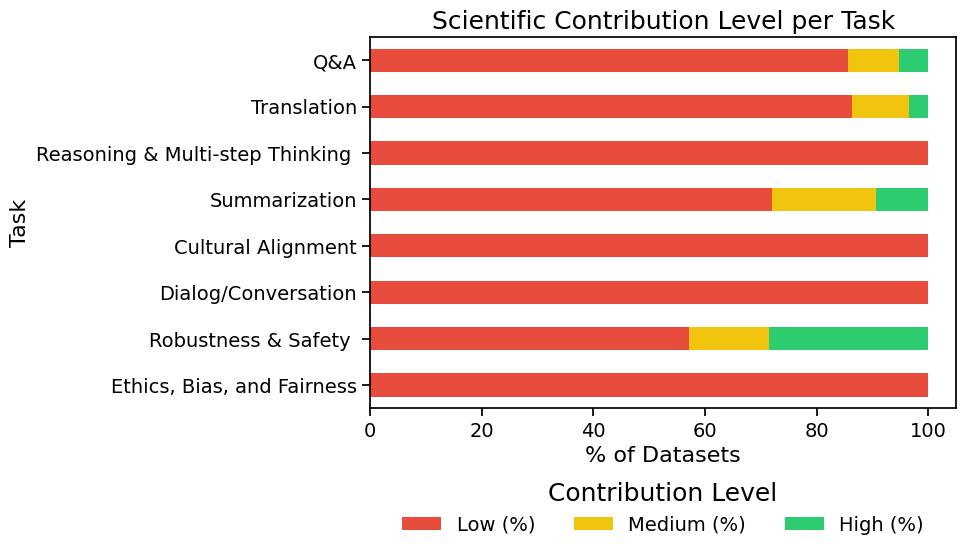}
    \caption{Scientific contribution level per task}
    \label{fig:scientific}
\end{subfigure}

\label{fig:quality_summary}
\caption{Overview of dataset quality across tasks. The subfigures present quality indicators including documentation, popularity, adoption, recency, licensing transparency, and scientific contribution. While the full taxonomy includes 12 tasks, we report results for the 9 tasks with available datasets.  \textit{Persona \& System Prompts}, and \textit{Function Call}, \textit{Code Generation}, and \textit{Official Documentation} are excluded as no datasets were available for those tasks.}
\label{fig:quality_summary}
\end{figure*}

\subsection{Dataset Results}
As shown in Figure~\ref{fig:dataset_distribution} and detailed in Appendix~\ref{appendix. stat_summary}, the distribution of the datasets is highly skewed towards specific tasks. For example, \textit{Translation} and \textit{Question Answering (Q\&A)} dominate, comprising 42.3\% and 38.3\% of the datasets, respectively. \textit{Summarization} adds another 12.3\%, while the remaining six tasks account for fewer than 30 datasets combined. Notably, \textit{Function Call}, \textit{Persona Ownership}, \textit{Code Generation}, and \textit{Official Documentation} have no datasets (zero datasets), revealing major gaps in current publicly available Arabic post-training resources.

\subsection{Automated Evaluation Results}

We present our findings from the automated evaluation of the collected datasets, focusing on their documentation and annotation quality, popularity, practical adoption, recency maintenance, licensing transparency, and scientific contribution, with detailed results shown in Appendix~\ref{appendix. quality_score_summary}.

\begin{itemize}[itemsep=0pt,topsep=0pt]
    \item \textbf{Documentation Quality:} Documentation standards show mixed results across tasks. Figure~\ref{fig:doc_dist} demonstrates that specialized domains like \textit{Ethics, Bias, and Fairness} and \textit{Robustness \& Safety} achieve excellent documentation quality (100\% high-quality scores). Still, these domains contain only 9 datasets in total, which may not adequately represent the broader landscape and could limit their applicability to diverse research contexts.
    
    \item \textbf{Popularity:} Dataset popularity varies significantly across tasks. Figure~\ref{fig:popularity} shows that traditional NLP tasks, like \textit{Q\&A}, \textit{Translation}, and \textit{Summarization}, include many widely-used datasets with strong community adoption. In contrast, tasks such as \textit{Dialog/Conversation} and \textit{Ethics, Bias, and Fairness} are dominated by low-popularity and medium-popularity datasets, reflecting either niche applications or limited awareness in the broader community.

    \item \textbf{Community Adoption:} Figure~\ref{fig:adoption} reveals consistently low adoption rates across all task categories, indicating limited reuse and citation of existing datasets. This pattern suggests that researchers may be creating new datasets rather than building upon existing work, potentially leading to fragmented efforts and reduced cumulative progress in the field.
    
    \item \textbf{Dataset Maintenance:} Maintenance practices vary considerably, highlighting inconsistent update schedules across the ecosystem. Figure~\ref{fig:recency} shows that newer research areas like \textit{Robustness \& Safety} and \textit{Ethics, Bias, and Fairness} maintain current datasets, while established tasks such as \textit{Summarization} and \textit{Translation} contain many outdated resources that lack regular maintenance cycles.
    
    \item \textbf{Licensing Transparency:} Licensing practices show positive trends toward open accessibility. Figure~\ref{fig:licencing} demonstrates that most Arabic datasets provide clear licensing information, with many adopting permissive licenses like Apache-2.0. This transparency facilitates both academic research and commercial applications, supporting broader utilization of Arabic post-training datasets.
    
    \item \textbf{Scientific Contribution:} Research integration remains limited across the dataset landscape. Figure~\ref{fig:scientific} indicates that most datasets lack formal scientific validation through peer-reviewed publications or DOI assignment. This gap suggests that many datasets represent individual contributions rather than systematically validated research contributions.
\end{itemize}

\section{Identified Gaps}

Our analysis has identified several critical gaps that could significantly hinder Arabic NLP research and applications (as per Table~\ref{tab:dataset-summary}). Potential gaps and limitations include the following:

\begin{itemize}[itemsep=0pt,topsep=0pt]
    \item \textbf{Limited Arabic Post-Training Data:} Small coverage of Arabic post-training datasets leads to slow advancement in Arabic-centric LLMs, and hence their applications. There are almost no Arabic datasets available for key post-training tasks such as \textit{Function Call}, \textit{Persona Ownership}, \textit{Code Generation}, and \textit{Official Documentation}. Undoubtedly, this scarcity significantly hampers the development of sophisticated Arabic large language models that can perform complex tasks.
    
    \item \textbf{Poor Dataset Documentation:} Poor documentation and annotation of datasets leads to invisible and inaccessible resources within the Arabic NLP community. As shown in Table~\ref{tab:table}, many valuable datasets remain uncategorized and difficult to discover, creating barriers for researchers who could benefit from existing work. This lack of proper documentation surely prevents the efficient reuse and building upon previous efforts in the field.

\begin{table}
  \centering\small
    \caption{Total of Arabic datasets categorized under the 12 selected tasks, compared to uncategorized datasets.}
  \begin{adjustbox}{width=0.47\textwidth}
    \begin{tabular}{lr}
      \toprule
      \textbf{Dataset Type} 
        & \textbf{Total} \\
      \midrule
      Categorized Datasets (for the 12 tasks) 
        & 366 \\
      Uncategorized Datasets 
        & 341 \\
      \bottomrule
    \end{tabular}
  \end{adjustbox}
  \label{tab:table}
  \vspace{-6pt}
\end{table}

    \item \textbf{Low Community Engagement:} Low popularity of Arabic datasets reflects how the Arabic NLP community remains small and sometimes discouraging to new contributors. This limited engagement raises research ethical issues, including failure to cite others' work and not giving proper credit to dataset creators. 
    
    \item \textbf{Limited Open-Source Integration:} Limited adoption of Arabic datasets in training open-source models and public Hugging Face spaces restricts the broader accessibility of Arabic NLP applications. One possible reason for this limited integration is the lack of computational resources available to researchers and practitioners working with Arabic language models. This creates a barrier that prevents the wider deployment and testing of Arabic NLP solutions in real-world applications.
    
    \item \textbf{Lack of Dataset Maintenance:} Lack of recency and maintenance characterizes the majority of Arabic datasets, with most open-source resources rarely receiving updates or maintenance for periods exceeding 12 months. This stagnation means that datasets become outdated and potentially less relevant to current research needs. The absence of regular updates suggests a lack of sustained community support and ongoing development efforts.
    
    \item \textbf{Weak Scientific Standards:} Weak scientific contribution characterizes most Arabic datasets, with almost all datasets not being released as part of peer-reviewed research papers or having DOI identifiers. The majority represent individual contributions rather than rigorous academic work, which typically results in lower quality standards. This pattern reflects poorly on the overall quality of Arabic datasets, as those released with research papers or DOIs tend to demonstrate higher quality and more thorough validation.
\end{itemize}

\begin{table*}[htp]
\centering
\small
\caption{Summary of Arabic Post-training Dataset Coverage and Key Identified Gaps}
\label{tab:dataset-summary}
\begin{tabular}{@{}lll@{}}
\toprule
\textbf{Category} & \textbf{Coverage} & \textbf{Key Gaps} \\ 
\midrule
Question Answering (Q\&A) & Strong (140 datasets) & Lacks community adoption \& scientific validation \\[2pt]
Translation & Strong (155 datasets) & Lacks community adoption \& needs maintenance \\[2pt]
Reasoning \& Multi-Step Thinking & Very limited (8 datasets) & Needs significant scale expansion \\[2pt]
Summarization & Moderate (45 datasets) & Lacks community adoption \& scientific rigor \\[2pt]
Cultural Alignment & Critically limited (3 datasets) &  Needs culturally nuanced datasets \\[2pt]
Dialog/Conversation & Very limited (6 datasets) & Lacks popularity \& needs maintenance \\[2pt]
Persona/Ownership/System Prompt & \textbf{No datasets} & Requires development \\[2pt]
Robustness \& Safety & Limited (8 datasets) & Needs broader coverage \& adoption \\[2pt]
Function Call & \textbf{No datasets} & Requires development \\[2pt]
Ethics, Bias, and Fairness & Critically limited (1 dataset) & Needs coverage \& licensing transparency \\[2pt]
Code Generation & \textbf{No datasets} & Requires development \\[2pt]
Official Documentation & \textbf{No datasets} & Requires development \\
\bottomrule
\end{tabular}
\end{table*}

\section{Case Study: Safety and Cultural Alignment }
Safety and cultural alignment datasets are crucial for developing responsible, culturally sensitive NLP systems. 
However, our findings reveal significant gaps in both areas. As shown in Figures~\ref{fig:dataset_distribution} and~\ref{fig:dataset_size_ranges}, \textit{Cultural Alignment} accounts for less than 1\% of all surveyed datasets, while \textit{Robustness \& Safety} includes only 8 datasets, with substantial variation in size and coverage. Both categories show consistently low adoption rates, and \textit{Cultural Alignment} additionally exhibits limited scientific contribution (Figure~\ref{fig:quality_summary}), suggesting underutilization despite the relatively strong popularity of some datasets.

This underrepresentation is especially concerning given the importance of cultural sensitivity and safety in Arabic-speaking contexts, where linguistic, societal, and religious norms differ greatly from dominant English-based benchmarks. The lack of culturally aware and safety-focused datasets increases the risk of deploying misaligned or even harmful NLP systems, like LLMs. To address these blind spots, we strongly recommend prioritizing the development of high-quality datasets tailored to Arabic cultural contexts and safety concerns, ensuring that future models are not only technically robust but also ethically and socially aligned.

\section{Recommendations and Future Directions}

The findings of this review highlight the strategic importance of post-training datasets for advancing Arabic-centric LLMs. While the existing resources on Hugging Face Hub provide a starting point, they fall short in coverage, documentation quality, cultural alignment, and scientific rigor. To address these limitations and accelerate the development of Arabic LLMs, we offer the following forward-looking recommendations, structured around priority domains, practical dataset creation strategies, and principles for collaborative research.

\subsection{High-Priority Domains for Future Post-Training Datasets}

This subsection outlines specific domains in Arabic post-training that are currently underrepresented or entirely missing, yet are crucial for building capable, safe, and culturally aligned Arabic LLMs. These domains should be prioritized in future post-training dataset development initiatives due to their strategic importance and lack of coverage.

\begin{itemize}[itemsep=0pt,topsep=0pt]
    \item \textit{Reasoning and Multi-Step Thinking:} Datasets supporting logical reasoning, problem-solving, and chain-of-thought prompting are vital for advanced LLM capabilities.

    \item \textit{Summarization:} While moderately covered, many existing datasets lack consistency in documentation, linguistic variety, and practical relevance to real-world 
    use cases.

    \item \textit{Cultural Alignment:} Data that reflects nuanced Arab world values, norms, and social constructs is crucial for building culturally sensitive NLP systems and applications.

    \item \textit{Dialog/Conversation:} This domain suffers from very limited coverage and low-quality documentation and annotation. Rich, dialect-sensitive dialogue datasets are essential for improving conversational fluency and natural interaction in Arabic-centric LLMs.

    \item \textit{Persona and System Prompting:} Needed for conversational agents to maintain consistent behavior and alignment across interactions.

    \item \textit{Robustness \& Safety:} Despite its importance for responsible AI development, the availability of high-quality Arabic post-training datasets in this domain remains limited.
    
    \item \textit{Function Calling:} Essential for tool-augmented NLP and API-connected LLMs, yet currently nonexistent in public Arabic post-training resources.
    
    \item \textit{Ethics, Bias, and Fairness:} Arabic datasets in this area are extremely limited, despite growing ethical concerns in global LLM adoption, development, and deployment.
    
    \item \textit{Code Generation:} There are currently no open Arabic datasets supporting code generation.

    \item \textit{Official Documentation:} This domain is completely absent from current post-training resources, although critical for building capable LLMs that can handle policies, manuals, formal content, or structured instructions.

\end{itemize}

\subsection{Practical Guidelines for Building Arabic Post-Training Datasets}

This subsection focuses on practical and scalable methods for creating Arabic post-training datasets. These guidelines are intended for researchers and developers, who aim to build new resources and address domain-specific gaps. The listed methods are grounded in existing tools, community collaboration, and modern data generation strategies.

\vspace{-3pt}

\paragraph{Dialectal Dialogue Collection}
Capturing authentic spoken Arabic from various dialect regions is essential. We recommend collecting spontaneous conversations from native speakers across the Arab world, followed by accurate transcription that preserves dialectal features.

\vspace{-3pt}

\paragraph{Collaborative Annotation Platforms}
A crowd-sourced annotation platform can empower native speakers to label data along cultural and contextual dimensions. By providing well-defined annotation guidelines, especially on culturally sensitive topics, the platform can produce high-quality datasets with rich sociocultural nuance.

\vspace{-3pt}

\paragraph{Human--LLM Hybrid Annotation}
Large language models can be leveraged to perform initial annotations, which are then verified or refined by human annotators. This semi-automated approach balances efficiency with quality assurance and reduces manual annotation overhead.

\paragraph{Synthetic Data Generation}
Arabic-capable LLMs can be prompted to generate new post-training data for underrepresented tasks. Although synthetic data offers scalability, rigorous validation is necessary to ensure linguistic correctness, cultural appropriateness, and task alignment.

\subsection{Recommendations for Future Research and Collaboration}

This final subsection presents high-level, strategic guidance for the broader research community. These recommendations emphasize principles like authenticity, cultural representation, and open collaboration. They are intended to shape future initiatives and encourage ethical, inclusive, and sustainable development of Arabic post-training datasets.

\begin{itemize}[itemsep=3pt,topsep=3pt]
        \item \textbf{Prioritize Missing Domains:} Direct funding, research, and community efforts toward domains with little to no coverage in Arabic (e.g., \textit{Function Calling} and \textit{Code Generation}).
    
    \item \textbf{Promote Authenticity over Translation:} Native Arabic content should be favored to avoid loss of context, nuance, or cultural misalignment present in translated material. While translated datasets can serve as a temporary bridge to address data scarcity, they fundamentally compromise the linguistic and cultural integrity essential for powerful Arabic LLMs. Native Arabic content preserves cultural subtleties, idiomatic expressions, and the language's unique morphological complexity that translation inevitably distorts. In culturally sensitive domains—including religious discourse, legal frameworks, and social interactions—native content ensures terminological accuracy and cultural appropriateness that directly impacts model performance and user acceptance. Thus, we recommend prioritizing investment in native Arabic dataset creation as a sustainable strategy for developing LLMs that authentically serve Arabic-speaking communities rather than imposing linguistic patterns from other language contexts.
    
    \item \textbf{Incorporate Cultural Context:} Datasets should reflect ethical, religious, and societal views, values, and cultures of the Arab world to ensure cultural robustness in AI outputs.
    
    \item \textbf{Broaden Linguistic Representation:} Both Modern Standard Arabic (MSA) and regional Dialectal Arabic (DA) should be represented in future dataset development to support real-world use cases across the Arab region.
    
    \item \textbf{Foster Open Collaboration and Transparency:} Dataset creators are encouraged to share licensing details, evaluation metrics, and use-case documentation to increase reproducibility, transparency, and adoption.

    \item \textbf{Investigate Dataset-Performance Relationships:} Future research should investigate relationships between our categorized dataset characteristics and actual model performance. Such studies could leverage our framework to conduct controlled experiments across task categories, establishing empirical relationships between dataset quality metrics and model effectiveness. This would provide valuable guidance for dataset creators and model developers in the Arabic NLP community.

\end{itemize}

\section{Conclusion}
In this paper, we conducted the first systematic survey of publicly available Arabic post-training datasets hosted on the Hugging Face Hub, with a focus on evaluating their quality, coverage, licensing transparency, and scientific contribution, across 12 key LLM capabilities. Our findings reveal several critical gaps, most notably the near absence of datasets in high-impact domains, such as \textit{Function Calling}, \textit{Code Generation}, \textit{Ethical Alignment}, and \textit{Official Documentation}. Despite the growing importance of post-training in aligning LLMs with human intent, Arabic remains substantially underrepresented in this space. Many existing datasets suffer from limited documentation, outdated maintenance, and low practical adoption. These shortcomings hinder the advancement of robust, culturally aligned, and ethically grounded Arabic LLMs.

We proposed a set of high-priority domains that require urgent dataset development and provided practical, scalable guidelines for building Arabic post-training resources through community collaboration, hybrid human–LLM annotation, and synthetic data generation. Additionally, we outlined strategic recommendations for promoting native content, cultural awareness, and linguistic diversity in future dataset creation efforts. Lastly, we released two open-source demo versions of our dataset collection and evaluation tools to the Arabic NLP research community. The introduction of these tools will facilitate standardized evaluation practices as well as reproducible research. In the near future, we aim to publicly share production versions with detailed documentation to ensure broad accessibility and adoption across research institutions.

\section*{Limitations}
While this study provides the first structured review of Arabic post-training datasets, it is subject to several limitations. First, this review covers only datasets openly available on Hugging Face Hub, omitting any private or gated resources.

Second, our collection and evaluation rely heavily on metadata and Dataset Cards (README) documentation, which may not always accurately reflect the actual quality or usability of the datasets. Some datasets may be underdocumented despite being high-quality in practice, and others may appear polished but lack effective downstream utility.

Third, this study does not assess how the reviewed datasets directly impact model performance. While our review provides essential infrastructure for dataset discovery, examining correlations between dataset characteristics and model effectiveness would require extensive computational resources and standardized benchmarking protocols beyond this study's scope. As such, the current study did not examine the relationship between the reviewed datasets and model performance.

\section*{Ethical Considerations}

While this study does not collect new data or generate text, analyzing public Arabic datasets raises ethical concerns, including unclear licensing, cultural bias, and dual-use risks. We encourage transparent licensing, inclusive annotations, and responsible governance in future dataset development.

\bibliography{custom}

\clearpage
\appendix

\onecolumn
\section{Evaluation Criteria}\label{appendix. criteria}

Appendix A presents a comprehensive scoring rubric for evaluating Arabic datasets across five key categories: Popularity, Adoption, Recency and Maintenance, Licensing Transparency, and Scientific Contribution, as shown in Table~\ref{tab:scoring-rubric}. Each category includes specific criteria and is scored based on defined numerical thresholds, which are then mapped to qualitative levels—High, Medium, or Low. For example, Popularity is measured by the number of likes and downloads, with a dataset considered highly popular if it receives a total of 200 or more. Adoption reflects how widely the dataset is used across models and spaces, while Recency and Maintenance assess how recently the dataset has been updated, rewarding more actively maintained resources.

Licensing Transparency evaluates whether the dataset includes a clear license, with high scores given to those that explicitly state a recognized license. In contrast, datasets marked as “unknown,” “other,” or “none” receive lower scores. The Scientific Contribution category assesses the dataset’s presence in the academic field, based on references to 
or arXiv papers and the inclusion of DOI objects. This rubric offers a structured framework for evaluating dataset quality and academic relevance, making it easier to compare datasets and identify those best suited for research and development in Arabic NLP.

\setlength{\tabcolsep}{10pt}
\renewcommand{\arraystretch}{1.3}


\begin{table*}[!ht]
\caption{Scoring rubric for evaluating Arabic datasets based on popularity, adoption, recency and maintenance, licensing transparency, and scientific contribution. Each criterion is scored individually and mapped to a qualitative level (High, Medium, or Low). The documentation criteria and scoring rubric are previously displayed in Table \ref{tab:tableA}.}
  \centering\small
  \begin{adjustbox}{width=\textwidth}
  \begin{tabular}{
    @{}p{0.18\textwidth}  
         p{0.3\textwidth}  
         p{0.15\textwidth}  
         p{0.20\textwidth}  
         p{0.13\textwidth}@{}
  }
    \toprule
    \textbf{Evaluation}
      & \textbf{Criteria}
      & \textbf{Score}
      & \textbf{Total Score}
      & \textbf{Level} \\
    \midrule

    \multirow{2}{*}{Popularity}
      & Dataset’s Number of Likes
      & Number of Likes
      & \multirow{2}{*}{%
          \makecell[l]{%
            $200 \le \mathrm{Score}$\\
            $100 \le \mathrm{Score} < 200$\\
            $\mathrm{Score} < 100$
          }}
      & \multirow{2}{*}{%
          \makecell[l]{High\\Medium\\Low}} \\
    \cmidrule(lr){2-3}
      & Dataset’s Number of Downloads
      & Number of Downloads
      &  & \\     
    \midrule

    \multirow{2}{*}{Adoption}
      & Number of Used Models
      & Number of Models
      & \multirow{2}{*}{%
          \makecell[l]{%
            $50 \le \mathrm{Score}$\\
            $20 \le \mathrm{Score} < 50$\\
            $\mathrm{Score} < 20$
          }}
      & \multirow{2}{*}{%
          \makecell[l]{High\\Medium\\Low}} \\
    \cmidrule(lr){2-3}
      & Number of Used Spaces
      & Number of Spaces
      &  & \\
    \midrule

    \multirow{3}{*}{Recency \& Maintenance}
      & Dataset’s Last Modified Date
      & Last Modified – Collection Date
      & \multirow{3}{*}{%
          \makecell[l]{%
            $\mathrm{Score}\le6\,$Mo\\
            $6\,$Mo $<\mathrm{Score}\le12\,$Mo\\
            $\mathrm{Score}>12\,$Mo
          }}
      & \multirow{3}{*}{%
          \makecell[l]{High\\Medium\\Low}} \\
      & 
      & 
      &  & \\
    \midrule

    \multirow{2}{*}{Licensing Transparency}
      & Dataset card states the license
      & License Name
      & \multirow{2}{*}{%
          \makecell[l]{%
            Known license\\
            'unknown'/'other'\\
            'none'
          }}
      & \multirow{2}{*}{%
          \makecell[l]{High\\Medium\\Low}} \\
    \cmidrule(lr){2-3}
      & Metadata tags state the license
      & License Name
      &  & \\
    \midrule

    \multirow{3}{*}{Scientific Contribution}
      & Dataset card includes ACL Papers
      & ACL Papers
      & \multirow{3}{*}{%
          \makecell[l]{%
            $3 \le \mathrm{Score}$\\
            $1 \le \mathrm{Score} < 3$\\
            $\mathrm{Score}=0$
          }}
      & \multirow{3}{*}{%
          \makecell[l]{High\\Medium\\Low}} \\
    \cmidrule(lr){2-3}
      & Metadata tags include ArXiv Papers
      & ArXiv Papers
      &  & \\
    \cmidrule(lr){2-3}
      & Metadata tags include a DOI Object
      & DOI Object
      &  & \\
    \bottomrule
  \end{tabular}
  \end{adjustbox}
  
  \label{tab:scoring-rubric}
\end{table*}

\newpage
\onecolumn
\section{Dataset Characteristics by Task}\label{appendix. stat_summary}

This appendix provides a comprehensive overview of dataset characteristics and quality across Arabic post-training tasks. Table~\ref{tab:dataset_summary} summarizes key statistics for each task category, including the number of datasets, average Hugging Face likes, downloads, model usage, and citation counts in ACL and ArXiv papers. These metrics offer insight into dataset visibility, reuse, and scholarly contribution.

\begin{figure}[!ht]
    \centering
    \includegraphics[width=1.0\linewidth]{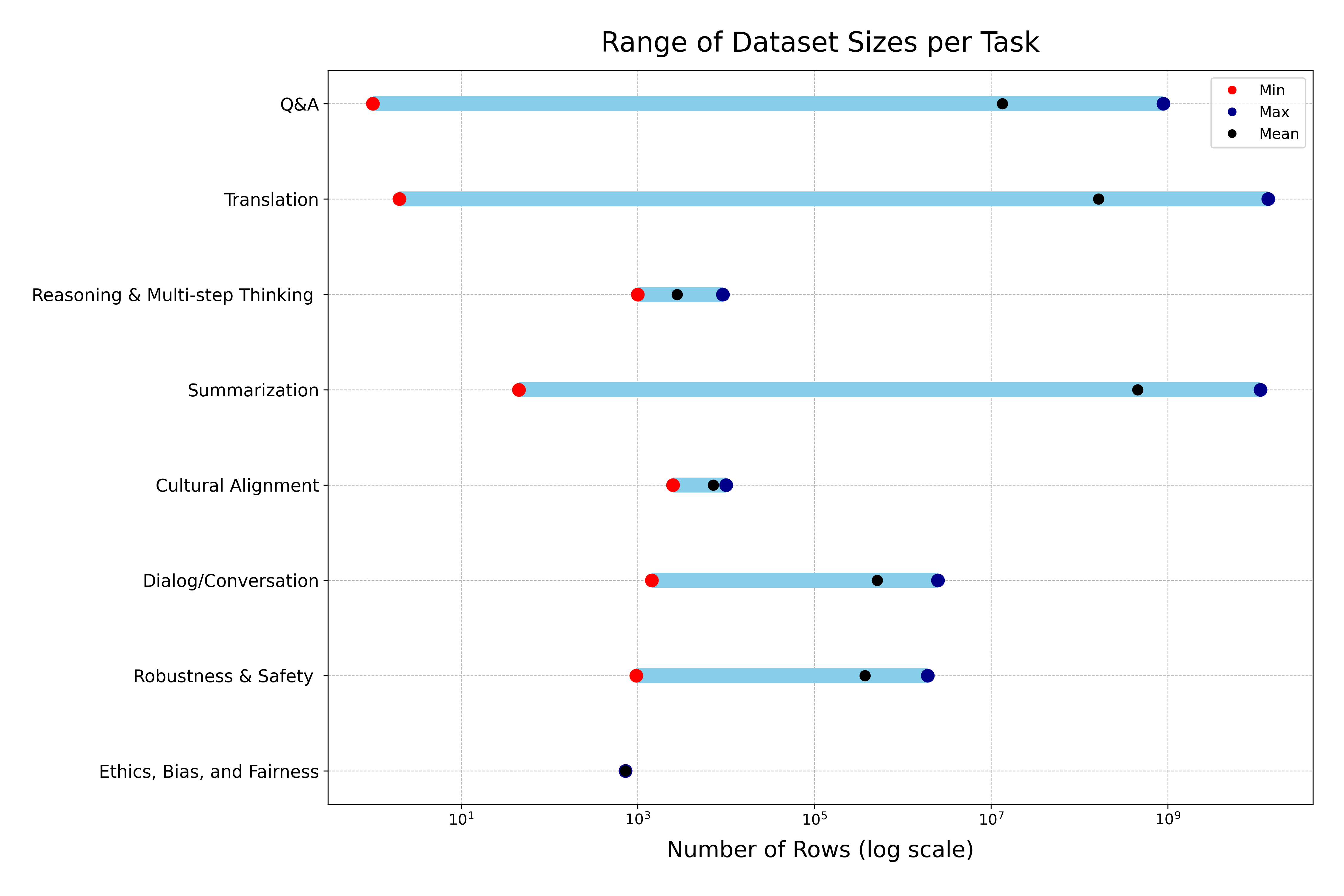}
    \caption{
        Range of dataset sizes per task (log scale). Each horizontal bar represents the minimum and maximum number of rows for datasets within a task, with red, blue, and black points denoting the minimum, maximum, and mean sizes, respectively. The wide variation in size highlights disparities in dataset availability and scale across post-training tasks. Although there are 12 tasks, here we only present the size of datasets with available data (n= 9). This figure reveals that dataset sizes vary dramatically not only across tasks but also within the same task category. Some tasks, such as Summarization and Translation, contain datasets ranging from a few dozen rows to over 10 billion. This high variance makes aggregate measures like the mean misleading; therefore, we emphasize range-based visualizations over summary statistics when discussing dataset scale.
    }
    \label{fig:dataset_size_ranges}
\end{figure}

Figure~\ref{fig:dataset_size_ranges} complements this summary by illustrating the range of dataset sizes per task on a logarithmic scale. This visualization reveals substantial variation both across and within tasks, with some datasets ranging from a few dozen to over 10 billion rows. Given this high variance, we emphasize range-based visualizations rather than relying solely on averages when assessing dataset scale.

\begin{table}[htbp]
\caption{
Values represent means with standard deviations in parentheses. For each task category, the table reports the number of datasets (n), mean number of Hugging Face likes and downloads, average count of model implementations, and mean number of ACL and ArXiv papers citing the dataset. For tasks with n = 1, standard deviations are not applicable and are indicated by (-). For tasks with n = 0, all values are indicated by (-) as no data is available.
}
\centering
\label{tab:dataset_summary}
\begin{adjustbox}{width=\textwidth}
\begin{tabular*}{\textwidth}{@{\extracolsep{\fill}}p{3cm}p{1.2cm}p{1.8cm}p{1.8cm}p{1.5cm}p{1.5cm}p{1.5cm}}
\hline
\textbf{Task} & \textbf{n} & \textbf{Likes} & \textbf{Downloads} & \textbf{Models} & \textbf{ACL  Papers} & \textbf{ArXiv Papers} \\
\hline
Q\&A & 140 & 10.6 (43.9) & 1285 (8288) & 3.1 (19.7) & 0.22 (0.61) & 0.27 (0.45) \\
\hline
Translation & 155 & 9 (20.5) & 721 (1805) & 1 (5.1) & 0.16 (0.52) & 0.21 (0.41) \\
\hline
Reasoning \& Multi-Step Thinking & 8 & 10 (11.6) & 105 (112) & 3.5 (7.2) & 0 (0) & 0 (0) \\
\hline
Summarization & 45 & 9.9 (22.9) & 2826 (13931) & 3 (12.6) & 0.33 (0.71) & 0.24 (0.43) \\
\hline
Cultural Alignment & 3 & 19.7 (27.4) & 171 (59) & 1.7 (1.5) & 0 (0) & 0.33 (0.58) \\
\hline
Dialog/Conversation & 6 & 1.8 (2.3) & 47 (42) & 0.2 (0.4) & 0 (0) & 0.17 (0.41) \\
\hline
Persona Ownership/System Prompt & 0 & - (-) & - (-) & - (-) & 0 (-) & 0 (-) \\
\hline
Robustness \& Safety & 8 & 4.9 (8.9) & 253 (167) & 1.4 (2.7) & 0.75 (1.04) & 0.62 (0.52) \\
\hline
Function Call & 0 & - (-) & - (-) & - (-) & 0 (-) & 0 (-) \\
\hline
Ethics, Bias, and Fairness & 1 & 16 (-) & 176 (-) & 0 (-) & 0 (-) & 0 (-) \\
\hline
Code Generation & 0 & - (-) & - (-) & - (-) & 0 (-) & 0 (-) \\
\hline
Official Documentation & 0 & - (-) & - (-) & - (-) & 0 (-) & 0 (-) \\
\hline
\end{tabular*}
\end{adjustbox}
\label{tab:dataset_summary}
\end{table}

\newpage

\newpage

\section{Quality Score Proportions By Task}
\label{appendix. quality_score_summary}
This appendix presents a task-level summary of dataset quality scores across six evaluation dimensions. Table~\ref{tab:task_quality_summary} reports the proportion of datasets rated as low, medium, or high for each criterion: documentation and annotation quality, popularity, adoption, recency and maintenance, licensing transparency, and scientific contribution. These scores reflect both the strengths and limitations of available Arabic post-training datasets and provide a quantitative basis for identifying quality gaps across task categories. Missing values are also reported to ensure transparency in coverage and support reproducibility. 

\begin{table*}[!ht]
\caption{
Dataset quality levels across tasks and evaluation dimensions. 
The \textbf{Missing} column refers to the number of datasets with missing scores for the specified level type.
For example, in the \textit{Robustness \& Safety} task, 2 datasets lack documentation level, and 1 lacks all evaluation scores. Tasks with no datasets are marked with (–).
}
\centering
\small
\begin{adjustbox}{width=0.95\textwidth,max height=0.8\textheight}
\begin{tabular}{lllrrrrr}
\toprule
\textbf{Task} & \textbf{\# Datasets} & \textbf{Missing} & \textbf{Level Type} & \textbf{Low (\%)} & \textbf{Medium (\%)} & \textbf{High (\%)} \\
\midrule
\multirow{6}{*}{Q\&A} 
& 140 & 8 & documentation\_annotation\_level & 3.79 & 46.21 & 50.00 \\
&  &  & popularity\_level & 58.33 & 13.64 & 28.03 \\
&  &  & adoption\_level & 97.73 & 0.76 & 1.52 \\
&  &  & recency\_maintenance\_level & 38.64 & 28.03 & 33.33 \\
&  &  & licensing\_transparency\_level & 11.36 & 4.55 & 84.09 \\
&  &  & scientific\_contribution\_level & 85.61 & 9.09 & 5.30 \\
\midrule
\multirow{6}{*}{Translation} 
& 155 & 9 & documentation\_annotation\_level & 4.79 & 47.95 & 47.26 \\
&  &  & popularity\_level & 56.85 & 19.18 & 23.97 \\
&  &  & adoption\_level & 100.00 & 0.00 & 0.00 \\
&  &  & recency\_maintenance\_level & 52.74 & 16.44 & 30.82 \\
&  &  & licensing\_transparency\_level & 22.60 & 13.70 & 63.70 \\
&  &  & scientific\_contribution\_level & 86.30 & 10.27 & 3.42 \\
\midrule
\multirow{6}{*}{Reasoning \& Multi-Step Thinking}
& 8 & 0 & documentation\_annotation\_level & 0.00 & 50.00 & 50.00 \\
&  &  & popularity\_level & 62.50 & 25.00 & 12.50 \\
&  &  & adoption\_level & 87.50 & 12.50 & 0.00 \\
&  &  & recency\_maintenance\_level & 0.00 & 37.50 & 62.50 \\
&  &  & licensing\_transparency\_level & 0.00 & 0.00 & 100.00 \\
&  &  & scientific\_contribution\_level & 100.00 & 0.00 & 0.00 \\
\midrule
\multirow{6}{*}{Summarization} 
& 45 & 2 & documentation\_annotation\_level & 6.98 & 39.53 & 53.49 \\
&  &  & popularity\_level & 60.47 & 9.30 & 30.23 \\
&  &  & adoption\_level & 97.67 & 0.00 & 2.33 \\
&  &  & recency\_maintenance\_level & 51.16 & 23.26 & 25.58 \\
&  &  & licensing\_transparency\_level & 18.60 & 4.65 & 76.74 \\
&  &  & scientific\_contribution\_level & 72.09 & 18.60 & 9.30 \\
\midrule
\multirow{6}{*}{Cultural Alignment}
& 3 & 0 & documentation\_annotation\_level & 0.00 & 66.67 & 33.33 \\
&  &  & popularity\_level & 0.00 & 66.67 & 33.33 \\
&  &  & adoption\_level & 100.00 & 0.00 & 0.00 \\
&  &  & recency\_maintenance\_level & 33.33 & 0.00 & 66.67 \\
&  &  & licensing\_transparency\_level & 33.33 & 0.00 & 66.67 \\
&  &  & scientific\_contribution\_level & 100.00 & 0.00 & 0.00 \\
\midrule
\multirow{6}{*}{Dialog/Conversation}
& 6 & 0 & documentation\_annotation\_level & 0.00 & 50.00 & 50.00 \\
&  &  & popularity\_level & 83.33 & 16.67 & 0.00 \\
&  &  & adoption\_level & 100.00 & 0.00 & 0.00 \\
&  &  & recency\_maintenance\_level & 66.67 & 0.00 & 33.33 \\
&  &  & licensing\_transparency\_level & 0.00 & 0.00 & 100.00 \\
&  &  & scientific\_contribution\_level & 100.00 & 0.00 & 0.00 \\
\midrule
\multirow{6}{*}{Robustness \& Safety}
& 8 & 2 & documentation\_annotation\_level & 0.00 & 0.00 & 100.00 \\
&  & 1 & popularity\_level & 14.29 & 28.57 & 57.14 \\
&  &  & adoption\_level & 100.00 & 0.00 & 0.00 \\
&  &  & recency\_maintenance\_level & 14.29 & 0.00 & 85.71 \\
&  &  & licensing\_transparency\_level & 0.00 & 0.00 & 100.00 \\
&  &  & scientific\_contribution\_level & 57.14 & 14.29 & 28.57 \\
\midrule
\multirow{6}{*}{Ethics, Bias, and Fairness}
& 1 & 0 & documentation\_annotation\_level & 0.00 & 0.00 & 100.00 \\
&  &  & popularity\_level & 0.00 & 100.00 & 0.00 \\
&  &  & adoption\_level & 100.00 & 0.00 & 0.00 \\
&  &  & recency\_maintenance\_level & 0.00 & 0.00 & 100.00 \\
&  &  & licensing\_transparency\_level & 100.00 & 0.00 & 0.00 \\
&  &  & scientific\_contribution\_level & 100.00 & 0.00 & 0.00 \\
\midrule
Persona Ownership/System Prompt & 0 & - & No data available & \multicolumn{3}{c}{-} \\
\midrule
Function Call & 0 & - & No data available & \multicolumn{3}{c}{-} \\
\midrule
Code Generation & 0 & - & No data available & \multicolumn{3}{c}{-} \\
\midrule
Official Documentation & 0 & - & No data available & \multicolumn{3}{c}{-} \\
\bottomrule
\end{tabular}
\end{adjustbox}
\label{tab:task_quality_summary}
\end{table*}

\end{document}

%% file: macros.tex
\definecolor{mypurple}{RGB}{111,61,121}